\documentclass[letterpaper]{article} 
\usepackage{aaai2026}  
\usepackage{times}  
\usepackage{helvet}  
\usepackage{courier}  
\usepackage[hyphens]{url}  
\usepackage{graphicx} 
\usepackage{natbib}  
\usepackage{caption} 
\usepackage{algorithm}
\usepackage{algorithmic}

\usepackage{multirow}
\usepackage{booktabs}
\usepackage{colortbl}
\usepackage{makecell}
\usepackage{pifont}
\usepackage{amsmath}
\usepackage{amssymb}

\usepackage{newfloat}
\usepackage{listings}
\DeclareCaptionStyle{ruled}{labelfont=normalfont,labelsep=colon,strut=off} 
\floatstyle{ruled}
\newfloat{listing}{tb}{lst}{}
\floatname{listing}{Listing}
\title{PriorRG: Prior-Guided Contrastive Pre-training and Coarse-to-Fine Decoding for Chest X-ray Report Generation}
\author{
    Kang Liu\textsuperscript{\rm 1,\rm 2,\rm3}, 
    Zhuoqi Ma\textsuperscript{\rm 1,\rm 2,\rm3}, 
    Zikang Fang\textsuperscript{\rm 1}, 
    Yunan Li\textsuperscript{\rm 1,\rm 2,\rm3}, 
    Kun Xie\textsuperscript{\rm 1,\rm 2,\rm3}, 
    Qiguang Miao\textsuperscript{\rm 1,\rm 2,\rm3}\thanks{Corresponding author}
}
\affiliations{
    \textsuperscript{\rm 1}School of Computer Science and Technology, Xidian University, Xi'an, Shaanxi 710071, China\\
    \textsuperscript{\rm 2}Xi'an Key Laboratory of Big Data and Intelligent Vision, Xi'an, Shaanxi 710071, China\\
    \textsuperscript{\rm 3}Key Laboratory of Collaborative Intelligence Systems, Ministry of Education, Xidian University, Xi'an 710071, China\\
    \{kangliu, 22009200766\}@stu.xidian.edu.cn, \{zhuoqima, yunanli, xiekun, qgmiao\}@xidian.edu.cn
}

\usepackage{bibentry}

\begin{document}

\maketitle

\begin{abstract}
Chest X-ray report generation aims to reduce radiologists' workload by automatically producing high-quality preliminary reports. A critical yet underexplored aspect of this task is the effective use of patient-specific prior knowledge---including clinical context (e.g., symptoms, medical history) and the most recent prior image---which radiologists routinely rely on for diagnostic reasoning. Most existing methods generate reports from single images, neglecting this essential prior information and thus failing to capture diagnostic intent or disease progression. To bridge this gap, we propose \textbf{PriorRG}, a novel chest X-ray report generation framework that emulates real-world clinical workflows via a two-stage training pipeline. In Stage 1, we introduce a prior-guided contrastive pre-training scheme that leverages clinical context to guide spatiotemporal feature extraction, allowing the model to align more closely with the intrinsic spatiotemporal semantics in radiology reports. In Stage 2, we present a prior-aware coarse-to-fine decoding for report generation that progressively integrates patient-specific prior knowledge with the vision encoder's hidden states. This decoding allows the model to align with diagnostic focus and track disease progression, thereby enhancing the clinical accuracy and fluency of the generated reports. Extensive experiments on MIMIC-CXR and MIMIC-ABN datasets demonstrate that PriorRG outperforms state-of-the-art methods, achieving a 3.6\% BLEU-4 and 3.8\% F1 score improvement on MIMIC-CXR, and a 5.9\% BLEU-1 gain on MIMIC-ABN.
\end{abstract}

\begin{links}
    \link{Code}{https://github.com/mk-runner/PriorRG}
    \link{Extended version}{https://arxiv.org/abs/2508.05353}
\end{links}

\section{Introduction}
\begin{figure}
    \centering
    \includegraphics[width=1\linewidth]{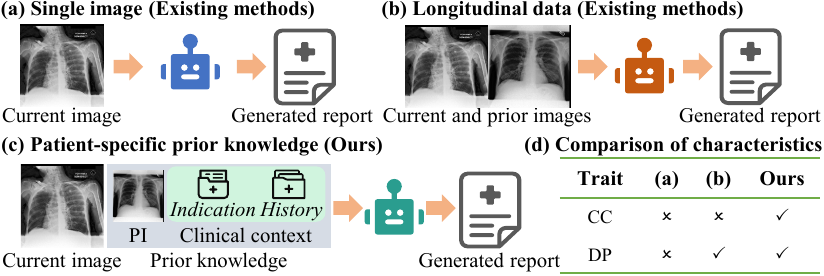}
    \caption{(a-c) show the workflow of existing methods and our approach. (d) summarizes their key properties. Longitudinal data includes both current and prior images. Patient-specific prior knowledge comprises the prior image (PI), \textit{indication}, and medical \textit{history}, which may be partially missing. ``CC'' and ``DP'' indicate whether a method models clinical context and disease progression, respectively.}
    \label{fig:introduce}
\end{figure}

Radiology report generation (RRG) \cite{aaai-2025-llm-rg4,2024-acmmm-sptio-temporal-fusion} leverages AI techniques to automatically interpret medical images---such as chest X-rays \cite{evoke,libra}, CT scans \cite{hamamci2024ct2rep,2025-ct-rrg-nature-communications,MicarVLMoE-2025}, and pathological slides \cite{guo2024histgen,2024-wsicaption-miccai}---and generate structured textual descriptions of clinically relevant findings. This automation supports radiologists by providing standardized preliminary reports, thereby improving diagnostic efficiency and consistency.

While recent RRG approaches have achieved remarkable progress, most models operate on isolated images (see Figure~\ref{fig:introduce}(a)) and overlook essential patient-specific prior knowledge, including both clinical context (i.e., the \textit{indication} and \textit{history} sections) and the most recent prior image. Such prior knowledge plays a critical role in real-world clinical reasoning by enabling personalized interpretation and tracking of disease progression. However, existing methods \cite{li-dcl,aaai-liu2024bootstrapping-llm,aaai-2025-mpo} largely ignore these factors, limiting their ability to produce context-aware, disease progression-oriented reports.

To capture the clinical context, \cite{ml4h-indication-rg} treats \textit{indications} as auxiliary input while ignoring the noise in the data. SEI \cite{sei} eliminates invalid or corrupted characters via preprocessing and employs a cross-modal fusion network to incorporate \textit{indications}, thereby generating accurate findings. Similarly, for \cite{2025-tmm-atl-ca}. However, they fail to consider longitudinal information, often resulting in hallucinations when describing disease progression. To combat this issue, existing methods incorporate prior images (Figure~\ref{fig:introduce}(b)) and employ techniques such as report pre-filling \cite{pre-fill-longitudinal-miccai-2023}, intra-modality similarity constraints \cite{aaai-2025-hc-llm}, and group causal transformers \cite{2024-eccv-hergen} to model temporal visual changes. Yet, they overlook clinical context, limiting their capacity to generate personalized and context-aware reports. This gap leads to our central research question: \textbf{Can we jointly model temporal visual changes and clinical context for improved cross-modal alignment and report generation?}

To address this challenge, we propose PriorRG, a novel chest X-ray report generation framework that mirrors real-world radiology workflows via a two-stage training pipeline. Stage~1 introduces a prior-guided contrastive pre-training scheme that simulates diagnostic reasoning by leveraging clinical context to guide spatiotemporal feature extraction, achieving better alignment with context-aware, disease progression-oriented radiology reports. Stage~2 presents a prior-aware coarse-to-fine decoding for report generation. We first devise an attention-enhanced layer fusion network to derive hierarchical visual representations from the vision encoder's hidden states. Motivated by principles of visual cognition, PriorRG progressively integrates clinical context, spatiotemporal information, and hierarchical visual cues in a coarse-to-fine manner. This design equips the report generator with rich, context-aware, disease progression-oriented representations, thereby enhancing both the clinical efficacy and linguistic quality of the generated reports. Comprehensive experiments on the MIMIC-CXR \cite{johnson-mimic-cxr-jpg} and MIMIC-ABN \cite{mimic-abn-ori} datasets demonstrate that our PriorRG significantly outperforms recent state-of-the-art methods in both medical image-text retrieval and radiology report generation. Our main contributions are:

\begin{itemize}
    \item We propose PriorRG, which integrates patient-specific prior knowledge to generate context-aware and disease progression-oriented reports.
    \item We introduce a prior-guided contrastive pre-training scheme that mirrors diagnostic reasoning by using clinical context to guide spatiotemporal features extraction, thereby improving alignment with report semantics and boosting medical image-text retrieval performance.
    \item We present a prior-aware coarse-to-fine decoding that incrementally integrates clinical context, disease progression patterns, and hierarchical visual cues, enhancing the clinical accuracy and fluency of generated reports.
\end{itemize}

\section{Related Work}
\textbf{Radiology report generation (RRG).} Unlike generic image captioning \cite{cvpr-2023conzic-image-captioning,2024-acmmm-image-caption}, RRG aims to generate detailed clinical descriptions for medical images \cite{mlrg}.  Recent advances have explored various techniques to improve accuracy, including knowledge graphs integration \cite{2025-kia-coling-knowledge}, contrastive learning \cite{sei,cofe-eccv-24}, retrieval-augmented methods \cite{jeong2024multimodal-MIDL,fse}, memory alignment \cite{chen-etal-2021-cross-modal,shen2024automatic_aaai}, human preference optimization \cite{2024-llm-cxr-rl,aaai-2025-mpo}, and LLM-based methods \cite{wang-2023-r2gengpt,aaai-liu2024bootstrapping-llm}. However, most existing methods rely solely on single-view images and overlook patient-specific prior knowledge---an essential component in real-world clinical decision-making. This limitation hinders their ability to capture clinical intent and monitor disease progression. To address this gap, we propose PriorRG that integrates prior knowledge via coarse-to-fine decoding to generate context-aware, disease progression-oriented reports.


\textbf{RRG via prior knowledge.} Several studies \cite{ml4h-indication-rg,sei,2025-tmm-atl-ca} leverage \textit{indications} to generate personalized reports, yet overlook disease progression---often leading to hallucinated descriptions of lesion changes. Recent efforts introduce longitudinal modeling via intra-modality similarity constraints \cite{aaai-2025-hc-llm}, report pre-filling \cite{pre-fill-longitudinal-miccai-2023}, or the group causal transformer \cite{2024-eccv-hergen} to capture temporal changes. While promising, they still underexploit individual clinical context, limiting their ability to infer patients' diagnostic intent. To bridge these gaps, we propose a prior-aware coarse-to-fine decoding that progressively integrates clinical context, disease progression patterns, and hierarchical visual cues, resulting in more accurate, fluent, and context-aware radiology reports.

\begin{figure*}
    \centering
    \includegraphics[width=1\linewidth]{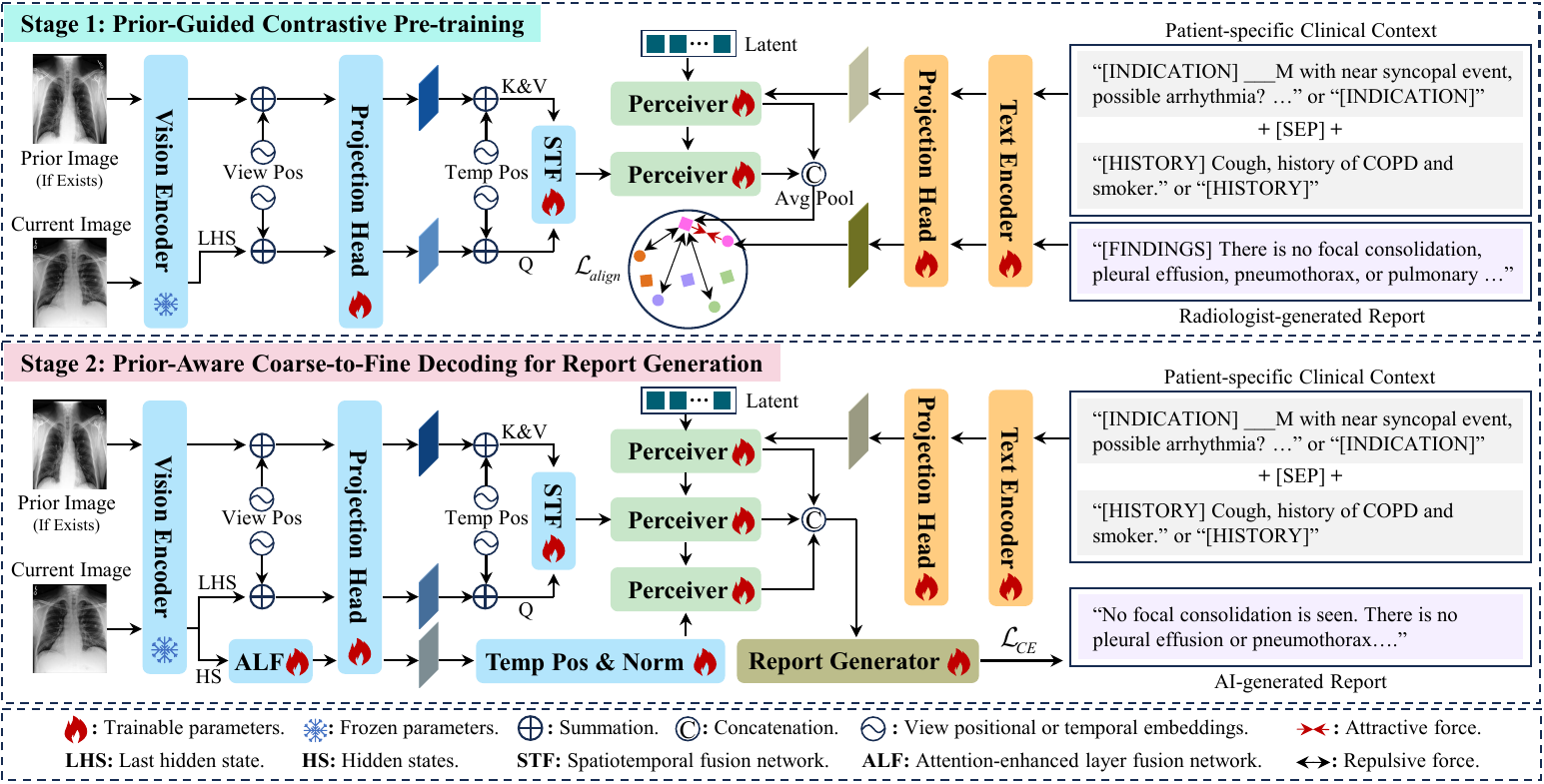}
    \caption{Overview of our PriorRG with a two-stage training pipeline, which consists of a vision encoder (RAD-DINO \cite{2024-rad-dino}), a text encoder (CXR-BERT \cite{2022-eccv-cxr-bert}), and a report generator (DistilGPT2 \cite{Sanh2019DistilBERTAD}).}
    \label{fig:overview}
\end{figure*}

\textbf{Medical vision-language pre-training (MVLP).} MVLP aims to learn joint visual-textual representations to support downstream tasks such as image-text retrieval \cite{medclip_wang_2022,zhang-2023-biomedclip,2023-cvpr-biovit-2301} and report generation \cite{promtmrg-aaai-2024,sei,mlrg}. GLoRIA \cite{Huang_2021_ICCV_gloria} and MGCA \cite{wang-mgca} enhance representations using multi-level cross-modal alignment, while SEI \cite{sei} links images to structured clinical entities. MedCLIP \cite{medclip_wang_2022} scales training via decoupled image-text matching based on the semantic similarity. To integrate domain knowledge, ARL \cite{chen2022align} and KAD \cite{zhang-kad} utilize UMLS for semantic alignment and zero-shot disease classification. Beyond static image-text pairs, \mbox{BioViL-T} \cite{2023-cvpr-biovit-2301} and MLRG \cite{mlrg} leverage longitudinal data to model disease progression. However, clinical context---crucial for diagnostic reasoning---remains largely underexplored. To address this, we introduce a prior-guided contrastive pre-training scheme that explicitly encodes clinical context and tracks disease progression, thereby improving cross-modal alignment.

\section{Method}

\subsection{Problem Statement}
Figure~\ref{fig:overview} illustrates the architecture of our PriorRG. The input includes a current image $x^{cur}_i$, a most recent prior image $x^{pri}_i$ (which may be absent), an indication $z_i$ (potentially missing), and a medical \textit{history} $h_i$ (possibly unavailable). We refer to $z_i$ and $h_i$ collectively as the clinical context for the $i^{th}$ sample. A key challenge lies in effectively leveraging this potentially incomplete prior information---$x^{pri}_i$, $z_i$, and $h_i$---to provide the model with personalized clinical background and disease progression cues. Our goal is to generate context-aware, disease progression-oriented radiology reports ${\hat{y}_i}$ by integrating all available inputs.


\subsection{Stage 1: Prior-Guided Contrastive Pre-training}
\textbf{Visual features extraction.} Following \cite{mlrg}, we employ the pre-trained vision encoder RAD-DINO \cite{2024-rad-dino} to extract visual features from chest X-rays. Radiographic view positions (e.g., AP vs. PA) significantly affect image appearance---for instance, the cardiac silhouette appears enlarged in AP views. To account for such projection-specific variations, we introduce a learnable view-position embedding and fuse it with the extracted visual features, enhancing the model's robustness to view-dependent discrepancies. The augmented features are then projected into a unified $d$-dimensional embedding space via a projection head. We denote the resulting visual features as ${\boldsymbol{V}} \in {\mathbb{R}^{M \times s \times d}}$, where $s$ is the sequence length, and $B$ represents batch size.

\textbf{Textual features extraction.} We adopt the pre-trained text encoder CXR-BERT \cite{2022-eccv-cxr-bert}, followed by a projection head. Specifically, the last hidden state from CXR-BERT is passed through the projection head to obtain textual features, denoted as ${\boldsymbol{T}} \in {\mathbb{R}^{B \times p \times d}}$, where $p$ is the number of tokens. To ensure consistency across different types of input, we prepend special tokens---``[INDICATION]'', ``[HISTORY]'', and ``[FINDINGS]''---to the \textit{indication}, medical \textit{history}, and radiology reports, respectively (see Figure \ref{fig:overview}). This design facilitates type-aware feature extraction while gracefully handling missing fields (e.g., absent \textit{indication} or medical \textit{history}), enabling a unified and robust encoding process.


\textbf{Spatiotemporal fusion network (STF).} While the current image alone can support report generation, relying solely on it may lead to hallucinations---particularly producing disease progression description (e.g., ``\textit{As compared to the previous radiograph, the patient has received a new right internal jugular vein catheter.}''). To mitigate this, we employ a ViT-style spatiotemporal fusion network \cite{mlrg,2021-vit} to model disease progression between current and prior images. The fusion process proceeds as follows: First, we inject temporal embeddings into both current and prior visual features to encode chronological relationships. Each STF block is then formulated as:
\begin{align}
{\boldsymbol{{V} }}^{st}_{ca} &= \text{LN}({\boldsymbol{V}}^{cur}+\text{CA}(\text{LN}({\boldsymbol{V}}^{cur}), \text{LN}({\boldsymbol{V}}^{pri}) )), \\
{\boldsymbol{{V} }}^{st} &= \text{LN}({\boldsymbol{{V} }}^{st}_{ca} + \text{FFN}({\boldsymbol{{V} }}^{st}_{ca}) ) ,
\end{align}
\noindent where ${\boldsymbol{V}}^{cur}$ and ${\boldsymbol{V}}^{pri}$ denote visual features of current and prior images, respectively. $\text{LN}(\cdot )$ and $\text{FFN}(\cdot)$ represent layer normalization and feed-forward network. $\text{CA}(\text{Q},\text{K\&V})$ denotes the cross-attention module used to capture inter-temporal dependencies. The number of STF blocks is empirically set to 3. For samples without a prior image, we directly treat ${\boldsymbol{V}}^{cur}$ as the spatiotemporal features ${\boldsymbol{{V} }}^{st} \in {\mathbb{R}^{B \times s \times d}}$.

\textbf{Instance-wise cross-modal alignment.} In clinical practice, physicians typically begin with an initial clinical assessment based on a patient's symptoms (i.e., \textit{indications}) and medical \textit{history}. Radiologists then integrate this clinical context with prior and current images to comprehensively evaluate lesion characteristics and progression. To simulate this diagnostic workflow, we incrementally incorporate clinical context ${\boldsymbol{T}}^{c}$ and spatiotemporal visual features ${\boldsymbol{\bar{V} }}^{st}$ using the Perceiver architecture \cite{icml-2021-perceiver}, formulated as: 
\begin{align}
\label{eq:percevior1}
{\boldsymbol{\bar{T} }}^{c} &= \text{Perceiver}({\boldsymbol{E}}^{lat}, {\boldsymbol{T}}^{c}), \\
\label{eq:percevior2}
{\boldsymbol{\bar{V}}}^{st} &= \text{Perceiver}
({\boldsymbol{\bar{T} }}^{c},{\boldsymbol{{V} }}^{st}), 
\end{align}
\noindent where $\text{Perceiver}({\boldsymbol{P}}, {\boldsymbol{Q}})$ is a modality-agnostic architecture that compresses input ${\boldsymbol{Q}}$ into a compact, learnable latent embedding ${\boldsymbol{P}}$ using the cross-attention module. Here, ${\boldsymbol{E}}^{lat} \in {\mathbb{R}^{B \times N \times d}}$ refers to the learnable latent embedding, where $N$ indicates the number of latents. ${\boldsymbol{\bar{T} }}^{c} \in {\mathbb{R}^{B \times N \times d}}$ denotes condensed clinical context feature, and ${\boldsymbol{\bar{V}}}^{st} \in {\mathbb{R}^{B \times N \times d}}$ signifies clinically informed spatiotemporal feature.

To ensure the consistency between image-report pairs, we employ the instance-wise cross-modal alignment \cite{wang-mgca,mlrg} to enhance multimodal representations. More concretely, we concatenate ${\boldsymbol{\bar{T} }}^{c}$ and ${\boldsymbol{\bar{V}}}^{st}$ along the sequence dimension, apply global average pooling, and perform L2 normalization to obtain global visual features ${\boldsymbol{V}}^{g} \in {\mathbb{R}^{B \times d}}$. The image-to-report similarity logits ${\boldsymbol{p}}^{i2r} \in {\mathbb{R}^{B \times B}}$ are then computed as:
\begin{align}
{\boldsymbol{p}}^{i2r}_{i}=\frac{\exp \left({\boldsymbol{V}}^{g}_i \cdot ({\boldsymbol{T}}^{g}_i)^T/\tau \right)}{ {\textstyle \sum_{j=1}^{B}\exp \left({\boldsymbol{V}}^{g}_i \cdot ({\boldsymbol{T}}^{g}_j)^T /\tau\right)} },
\end{align}
\noindent where ${\boldsymbol{T}}^{g}_i \in {\mathbb{R}^{B \times d}}$ denotes the global textual features for radiologist-generated reports. $\tau$ is the temperature parameter. Similarly, the report-to-image similarity logits are described as ${\boldsymbol{p}}^{r2i} \in {\mathbb{R}^{B \times B}}$. To account for the possibility of multi-view images during a visit, we treat all image-report pairs associated with the same visit as positive pairs, resulting in multiple positive pairs. Consequently, the ground-truth matching label matrix ${\boldsymbol{q}} \in {\mathbb{R}^{B \times B}}$ is defined as:
\begin{align}
{\boldsymbol{q}_{i,j}} = \frac{{{\mathbb{I}_{\text{equal} }}\left( {{y_{i}}, {y_{j}}} \right)}}{{\sum\nolimits_{k = 1}^{B} {{\mathbb{I}_{\text{equal} }}\left( {{y_{i}},{y_{k}}} \right)} }},
\end{align}
\noindent where ${\mathbb{I}_{\text{equal} }} ({{y_{i}}, {y_{j}}})$ is an indicator function that equals 1 if the $i^{th}$ and $j^{th}$ samples share the same report (i.e., $y_i=y_j$), and 0 otherwise. The instance-wise cross-modal alignment loss is defined as the cross-entropy loss between ground-truth matching label matrix ${\boldsymbol{q}}$ and similarity logits ${\boldsymbol{p}}$:
\begin{align}
{{{\mathcal{L}}}_{align}} =  - \frac{1}{2B}\sum\limits_{k = 1}^B {\left( {{\boldsymbol{q}}_k \log {\boldsymbol{p}}_k^{i2r} + {\boldsymbol{q}}_k\log {\boldsymbol{p}}_k^{r2i}} \right)}.
\end{align}
\noindent To summarize, Stage 1 optimizes the alignment loss ${{{\mathcal{L}}}_{align}}$, which leverages clinical context to guide the extraction of spatiotemporal features, enhancing cross-modal alignment and boosting medical image-text retrieval performance.

\begin{table*}
\centering
\setlength{\tabcolsep}{2mm}
\begin{tabular}{cccccccccccc} 
\toprule
\multirow{2}{*}{\textbf{Dataset}} & \multirow{2}{*}{\textbf{Method}} & \multirow{2}{*}{\textbf{Venue}} & \multicolumn{6}{c}{\textbf{NLG Metrics} $\uparrow$} & \multicolumn{3}{c}{\textbf{CE Metrics} $\uparrow$} \\
\cmidrule(r){4-9}\cmidrule(lr){10-12}
 &  &  & \textbf{B-1} & \textbf{B-2} & \textbf{B-3 }& \textbf{B-4} & \textbf{MTR} & \textbf{R-L} & \textbf{P} & \textbf{R} & \textbf{F1 }\\ 
\midrule
\multirow{13}{*}{M-CXR} & KiUT & CVPR'23 & 0.393 & 0.243 & 0.159 & 0.113 & 0.160 & 0.285 & 0.371 & 0.318 & 0.321 \\ 
 & METransformer & CVPR'23 & 0.386 & 0.250 & 0.169 & 0.124 & 0.152 & 0.291 & 0.364 & 0.309 & 0.311 \\ 
 & CoFE & ECCV'24 & - & - & - & 0.125 & \underline{0.176} & {0.304} & 0.489 & 0.370 & 0.405 \\ 
 & DCG & MM'24 & 0.397 & 0.258 & 0.166 & 0.126 & 0.162 & 0.295 & 0.441 & 0.414 & 0.404 \\ 
 & MAN & AAAI'24 & 0.396 & 0.244 & 0.162 & 0.115 & 0.151 & 0.274 & 0.411 & 0.398 & 0.389 \\ 
 & R2GenGPT & Meta-Radio'23 & {0.411} & {0.267} & {0.186} & 0.134 & 0.160 & 0.297 & 0.392 & 0.387 & 0.389 \\ 
 & Med-LLM & MM'24 & - & - & - & 0.128 & 0.161 & 0.289 & 0.412 & 0.373 & 0.395 \\ 
 & R2-LLM & AAAI'24 & 0.402 & 0.262 & 0.180 & 0.128 & 0.175 & 0.291 & 0.465 & \underline{0.482} & \underline{0.473} \\ 
 & SEI & MICCAI'24 & 0.382 & 0.247 & 0.177 & {0.135} & 0.158 & 0.299 & \underline{0.523} & 0.410 & 0.460 \\ 
 & HERGen & ECCV'24 & 0.395 & 0.248 & 0.169 & 0.122 & 0.156 & 0.285 & - & - & - \\ 
 & MPO & AAAI'25 & \textbf{0.416} & \underline{0.269} & \underline{0.191} & \underline{0.139} & 0.162 & \underline{0.309} & 0.436 & 0.376 & 0.353 \\ 
 & \textbf{PriorRG (Ours)} & - & \underline{0.412} & \textbf{0.290} & \textbf{0.220} & \textbf{0.175} & \textbf{0.189} & \textbf{0.324} & \textbf{0.541} & \textbf{0.485} & \textbf{0.511} \\
 & $\bigtriangleup(\%) \uparrow$ & - & -0.4 & +2.1 & +2.9 & +3.6 & +1.3 & +1.5 & +1.8 & +0.3 & +3.8 \\ 
\midrule
\multirow{4}{*}{M-ABN} & CMN$^\Diamond $ & ACL'21 & 0.256 & 0.147 & 0.095 & 0.066 & 0.110 & 0.230 & \underline{0.466} & \underline{0.454} & \underline{0.460} \\ 
 & SEI$^\Diamond$ & MICCAI'24 & \underline{0.267} & \underline{0.157} & \underline{0.104} & \underline{0.073} & \underline{0.114} & \underline{0.231} & \underline{0.466} & 0.408 & 0.435 \\ 
 & \textbf{PriorRG (Ours)} & - & \textbf{0.326} & \textbf{0.201} & \textbf{0.139} & \textbf{0.102} & \textbf{0.140} & \textbf{0.242} & \textbf{0.467} & \textbf{0.476} & \textbf{0.471} \\
 & $\bigtriangleup(\%) \uparrow$ & - & +5.9 & +4.4 & +3.5 & +2.9 & +2.6 & +1.1 & +0.1 & +2.2 & +1.1 \\
\bottomrule
\end{tabular}
\caption{Comparison of report generation performance with SOTA methods on MIMIC-CXR (M-CXR) and MIMIC-ABN (M-ABN) datasets. $\bigtriangleup$ indicates the performance difference between PriorRG and the best baseline. $\Diamond$ denotes reproduced results; others are cited from the original papers. The \textbf{best} and \underline{second-best} values are in \textbf{bold} and \underline{underlined}, respectively.}
\label{tab:main-result}
\end{table*}


\begin{figure}
    \centering
    \includegraphics[width=1\linewidth]{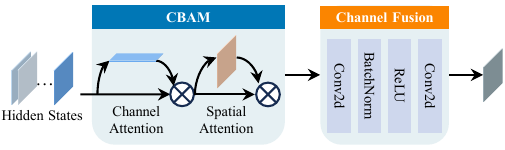}
    \caption{Overview of attention-enhanced layer fusion network (ALF), which leverages CBAM \cite{Woo_2018_ECCV-cbam} to extract hierarchical visual features.}
    \label{fig:stf-alf}
\end{figure}

\subsection{Stage 2: Prior-Aware Coarse-to-Fine Decoding for Report Generation}
Stage 2 consists of two main components: (1) an attention-enhanced hierarchical fusion network for constructing hierarchical visual semantics, and (2) a coarse-to-fine decoding that progressively integrates prior knowledge with hierarchical visual semantics to guide report generation.

\textbf{Attention-enhanced layer fusion network.} Previous methods \cite{liang2024divide-acmmm-24,mlrg} rely solely on the vision encoder's last hidden state, overlooking low-level details such as lesion morphology. To mitigate this, we design an attention-enhanced layer fusion network based on CBAM \cite{Woo_2018_ECCV-cbam} (see Figure~\ref{fig:stf-alf}). CBAM's channel and spatial attention are applied to each encoder layer to highlight diagnostically relevant features. The refined features are fused through a Conv2D projector that preserves both spatial and semantic information. A projection head with temporal positional embedding and layer normalization then yields hierarchical visual representations ${\boldsymbol{V}}^{hier} \in \mathbb{R}^{B \times s \times d}$ capturing multi-level cues.

\textbf{Prior-aware coarse-to-fine decoding.} Motivated by principles of visual cognition and radiologists' diagnostic workflow, we propose a prior-aware coarse-to-fine decoding. Specifically, we first extract the condensed clinical context feature ${\boldsymbol{\bar{T} }}^{c}$ and the clinically informed spatiotemporal feature ${\boldsymbol{\bar{V}}}^{st}$ via Equations~(\ref{eq:percevior1}) and~(\ref{eq:percevior2}). These features are derived from the last hidden state of the text and vision encoders, and thus encode high-level semantics of patient clinical background and disease progression. We treat them as coarse-grained priors that offer a holistic overview for guiding subsequent fine-level decoding. We then enhance ${\boldsymbol{\bar{V}}}^{st}$ with hierarchical visual representations ${\boldsymbol{{V} }}^{hier}$ using the Perceiver architecture \cite{icml-2021-perceiver}, formulated as:
\begin{align}
\label{eq:percevior3}
{\boldsymbol{\bar{V}}}^{hier} = \text{Perceiver}
({\boldsymbol{\bar{V}}}^{st},{\boldsymbol{{V} }}^{hier}).
\end{align}
The resulting ${\boldsymbol{\bar{V}}}^{hier}$ further refines and enriches fine-grained representations. Finally, we concatenate ${\boldsymbol{\bar{T} }}^{c}$, ${\boldsymbol{\bar{V}}}^{st}$, and ${\boldsymbol{\bar{V}}}^{hier}$ along the sequence dimension before feeding them into the report generator, thereby producing context-aware, disease progression-oriented reports. 

\textbf{Report generation.} We adopt the pre-trained DistilGPT2 \cite{Sanh2019DistilBERTAD} to generate free-text reports. Stage 2 is trained to minimize the cross-entropy loss between the AI-generated report ${\hat{y}_i}$ and the radiologist-generated report ${y_i}$. The objective function is formulated as:
\begin{align}
{\mathcal{{L}}}_{{CE}}^i=-\sum_{k=1}^{K} \log p({\hat y}_i^k \mid {\hat y}_i^{<k},x_i^{cur},x_i^{pri},z_i,h_i),
\end{align}
\noindent where $K$ denotes the maximum number of generated tokens. ${\hat y}_i^{<k}$ represents the tokens generated before step $k$.

\section{Experiments}
\subsection{Experimental Settings}
\textbf{Datasets.} \textbf{(1) MIMIC-CXR} \cite{johnson-mimic-cxr-jpg} is a large-scale, publicly available dataset of chest X-rays paired with free-text radiology reports. For each case, we organize data by ``study id'' and retrieve the most recent prior image when available. \textbf{(2) MIMIC-ABN} \cite{mimic-abn-ori} is a curated subset of MIMIC-CXR that focuses exclusively on abnormal findings in reports. Following previous studies \cite{chen-etal-2020-generating,sei,aaai-2025-mpo}, we treat only the ``Findings'' section as the radiologist-generated report, discarding empty or non-informative entries. All experiments adhere to the official dataset partitions. Dataset statistics are presented in Appendix Table \ref{tab:statistic}.

\textbf{Evaluation metrics.} Following protocols in \cite{wang2023metransformer,liang2024divide-acmmm-24,aaai-liu2024bootstrapping-llm}, we evaluate the quality of AI-generated reports using both natural language generation (NLG) and clinical efficacy (CE) metrics. For NLG metrics, which measure linguistic similarities between AI- and radiologist-generated reports, we use BLEU-n (B-n), METEOR (MTR), and ROUGE-L (R-L). For CE metrics, we assess clinical relevance by computing the micro-average Precision (P), Recall (R), and F1-score (F1) across 14 observations annotated by CheXpert \cite{irvin-chexpert}. 

\textbf{Implementation details.} The unified feature dimension is set to $d=768$, the number of latents to $N=128$, the maximum generation length to $K=100$ tokens, and the beam size to 3 during inference. We apply early stopping and learning rate scheduling for training stability. Additional implementation details are in Appendix Section A.

\subsection{Main Results}
\textbf{Baseline approaches.} We compare our PriorRG against 12 state-of-the-art (SOTA) methods, categorized into eight groups: knowledge graph-based methods (KiUT \cite{huang-kiut} and METransformer \cite{wang2023metransformer}), contrastive learning framework (CoFE \cite{cofe-eccv-24}), retrieval-augmented method (DCG \cite{liang2024divide-acmmm-24}), memory alignment approaches (CMN \cite{chen-etal-2021-cross-modal} and MAN \cite{shen2024automatic_aaai}), LLM-based approaches (R2GenGPT \cite{wang-2023-r2gengpt}, Med-LLM \cite{liu2024in-context-acmmm}, and R2-LLM \cite{aaai-liu2024bootstrapping-llm}), clinical context-driven model (SEI \cite{sei}), longitudinal data-based method (HERGen \cite{2024-eccv-hergen}), and human preference optimization (MPO \cite{aaai-2025-mpo}).

\begin{table*}
\centering
\setlength{\tabcolsep}{1.8mm}
\begin{tabular}{cccccccccccccc} 
\toprule
\multirow{2}{*}{\textbf{Model}} & \multirow{2}{*}{\textbf{Stage 1}} & \multicolumn{3}{c}{\textbf{Stage 2}} & \multicolumn{6}{c}{\textbf{NLG Metrics} $\uparrow$} & \multicolumn{3}{c}{\textbf{CE Metrics} $\uparrow$} \\ 
\cmidrule(r){3-5}\cmidrule(lr){6-11}\cmidrule(lr){12-14}
 &  & \textbf{CC} & \textbf{PI} & \textbf{Hidden States} & \textbf{B-1} & \textbf{B-2} & \textbf{B-3} & \textbf{B-4} & \textbf{MTR} & \textbf{R-L} & \textbf{P} & \textbf{R} & \textbf{F1} \\ 
\midrule
(a) & \ding{51} & \ding{55} & \ding{55} & \ding{55} & 0.355 & 0.219 & 0.149 & 0.108 & 0.154 & 0.262 & 0.521 & 0.431 & 0.472 \\
(b) & \ding{51} & \ding{55} & \ding{55} & \ding{51} & 0.357 & 0.216 & 0.143 & 0.102 & 0.154 & 0.255 & 0.513 & 0.480 & 0.496 \\
(c) & \ding{51} & \ding{51} & \ding{55} & \ding{55} & 0.405 & 0.283 & 0.214 & 0.170 & 0.186 & 0.320 & 0.517 & 0.460 & 0.487 \\
(d) & \ding{51} & \ding{51} & \ding{55} & \ding{51} & 0.404 & 0.285 & 0.217 & 0.173 & 0.187 & \textbf{0.328} & 0.540 & 0.477 & 0.507 \\
(e) & \ding{51} & \ding{51} & \ding{51} & \ding{55} & 0.400 & 0.282 & 0.214 & 0.171 & 0.186 & 0.323 & 0.539 & 0.465 & 0.499 \\
(f) & \ding{55} & \ding{51} & \ding{51} & \ding{51} & 0.387 & 0.271 & 0.206 & 0.165 & 0.180 & 0.320 & 0.526 & 0.408 & 0.459 \\
\textbf{PriorRG} & \ding{51} & \ding{51} & \ding{51} & \ding{51} & \textbf{0.412} & \textbf{0.290} & \textbf{0.220} & \textbf{0.175} & \textbf{0.189} & 0.324 & \textbf{0.541} & \textbf{0.485} & \textbf{0.511} \\
\bottomrule
\end{tabular}
\caption{Ablation study on MIMIC-CXR dataset. ``CC'' and ``PI'' denote clinical context and prior image, respectively.}
\label{tab:ablation-study}
\end{table*}

\textbf{Comparison with SOTA methods.} Table~\ref{tab:main-result} presents a comparison between our PriorRG and SOTA methods on MIMIC-CXR \cite{johnson-mimic-cxr-jpg} and MIMIC-ABN \cite{mimic-abn-ori} datasets. The results show that our PriorRG consistently surpasses existing methods in both linguistic quality and clinical accuracy. On the MIMIC-CXR dataset, PriorRG achieves a notable 3.6\% improvement in B-4, indicating better matching accuracy in longer n-gram sequences. Additionally, gains in MTR (+1.3\%) and R-L (+1.5\%) highlight improvements in lexical diversity, synonym matching, and semantic consistency. In terms of CE metrics, PriorRG boosts the F1 by 3.8\%, demonstrating greater reliability in clinical accuracy. On the MIMIC-ABN dataset, PriorRG achieves a 5.9\% increase in B-1 and a 1.1\% improvement in F1, emphasizing its effectiveness in describing abnormal findings. These improvements stem from our proposed prior-aware coarse-to-fine decoding, which enables the model to recognize individual clinical contexts and capture disease progression. 
Appendix presents extra results (Table \ref{tab:iu-xray}) and comparisons with MLRG (Tables \ref{tab:mlrg-input}--\ref{tab:mlrg-performance}).

\textbf{Clinical accuracy of 14 observations.} Appendix Table \ref{tab:chexbert-details} presents the clinical accuracy of 14 observations extracted by the CheXpert \cite{irvin-chexpert}. Our PriorRG outperforms the clinically-informed SEI model \cite{sei} on 13 of 14 observations in terms of F1-score, highlighting the superior clinical accuracy of generated reports.


\begin{figure}
    \centering
    \includegraphics[width=0.85\linewidth]{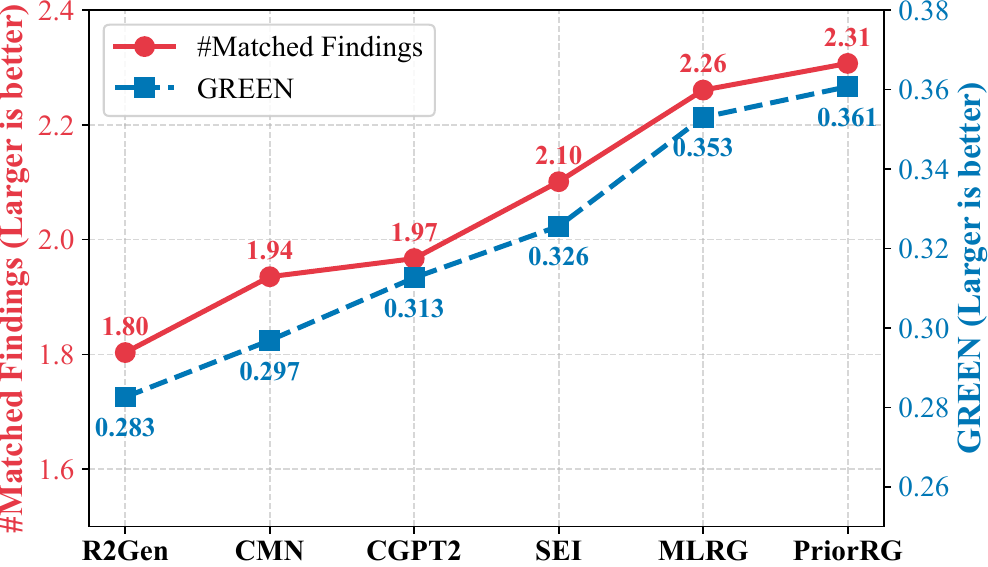}
    \caption{Expert-aligned assessment on the MIMIC-CXR dataset. ``\#Matched Findings'' indicates the average matched findings between AI- and radiologist-generated reports.}
    \label{fig:green}
\end{figure}


\textbf{Evaluation with large language model.} We assess clinical consistency and expert alignment of generated reports using two metrics: ``\#Matched Findings'' and the GREEN score \cite{2024-green}. The GREEN score combines clinical significance errors with ``\#Matched Findings,'' offering a robust evaluation aligned with expert preferences. Both metrics are derived via the pre-trained GREEN-RadLlama2-7B model. We benchmark PriorRG against CMN \cite{chen-etal-2021-cross-modal}, CGPT2 \cite{nicolson-improving-cvt2distilgpt2}, SEI \cite{sei}, and MLRG \cite{mlrg}. Figure~\ref{fig:green} shows that PriorRG significantly outperforms all baselines across both metrics, confirming its ability to generate clinically reliable, expert-aligned reports.

\subsection{Ablation Study}

\begin{figure}
    \centering
    \includegraphics[width=1\linewidth]{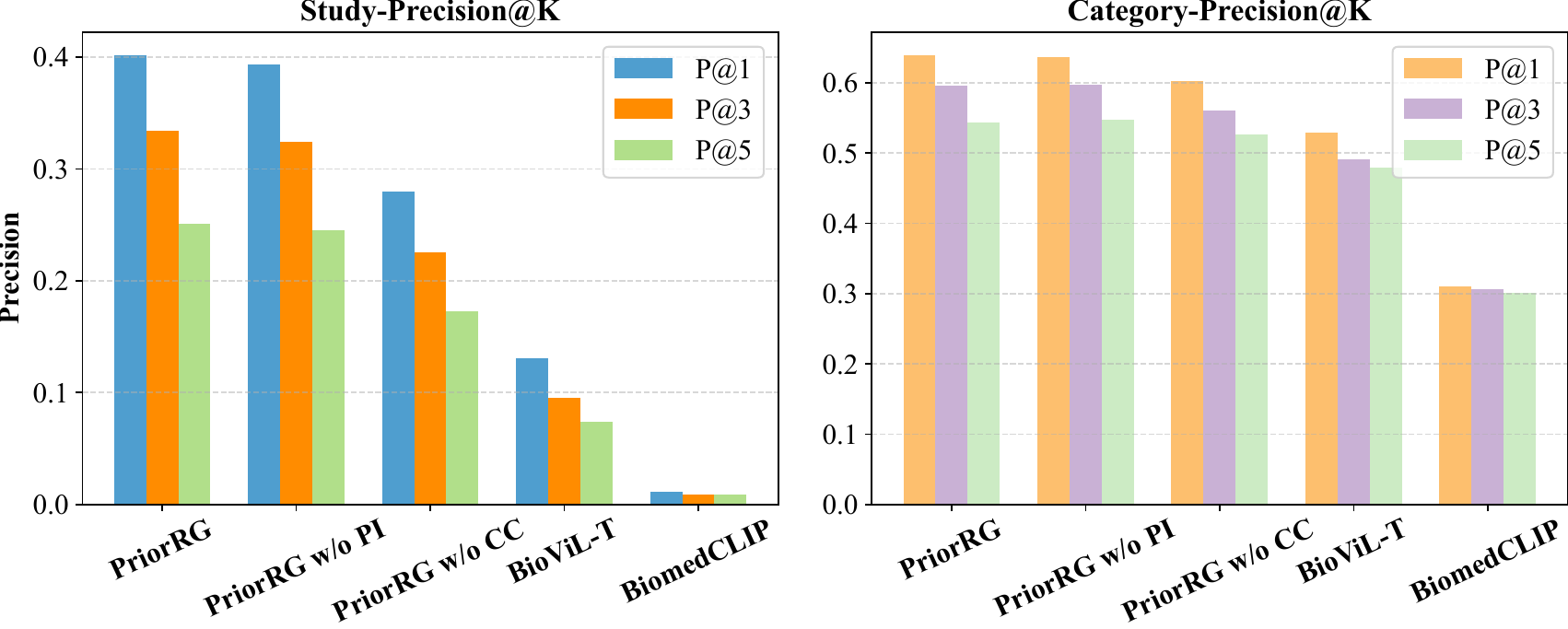}
    \caption{Medical image-text retrieval results on the MIMIC-5x200 dataset. ``PI'' and ``CC'' denote prior images and clinical context, respectively.}
    \label{fig:retrieval}
\end{figure}

\textbf{Effect of prior-guided contrastive pre-training (Stage 1).} We evaluate Stage 1 using a medical image-text retrieval task on the MIMIC-5x200 dataset \cite{johnson-mimic-cxr-jpg}. Following the protocols in \cite{pmlr-v182-zhang22a-5x200,nips-2024-benchx}, we construct the MIMIC-5x200 dataset by sampling 200 test set instances for each of five common diseases--- \textit{Atelectasis}, \textit{Cardiomegaly}, \textit{Consolidation}, \textit{Edema}, and \textit{Pleural Effusion}---from MIMIC-CXR, resulting in a total of 1,000 samples. Given a query image, we retrieve the top-K most similar reports from this dataset and evaluate performance using Category-Precision@K (Cat-P@K) and Study-Precision@K (Stu-P@K). Cat-P@K measures whether retrieved reports belong to the same disease category as the query image, while Stu-P@K evaluates whether retrieved reports originate from the same study. As illustrated in Figure~\ref{fig:retrieval}, PriorRG surpasses BiomedCLIP \cite{zhang-2023-biomedclip}, BioViL-T \cite{2023-cvpr-biovit-2301}, and ablated variants across both metrics. These results indicate that Stage 1 effectively encodes prior knowledge into semantically rich multimodal representations. Furthermore, Table~\ref{tab:ablation-study} shows that removing Stage 1 (variant (f)) leads to significant drops in both clinical accuracy and language quality, confirming its central role in improving downstream report generation.

\begin{figure*}
    \centering
    \includegraphics[width=1\linewidth]{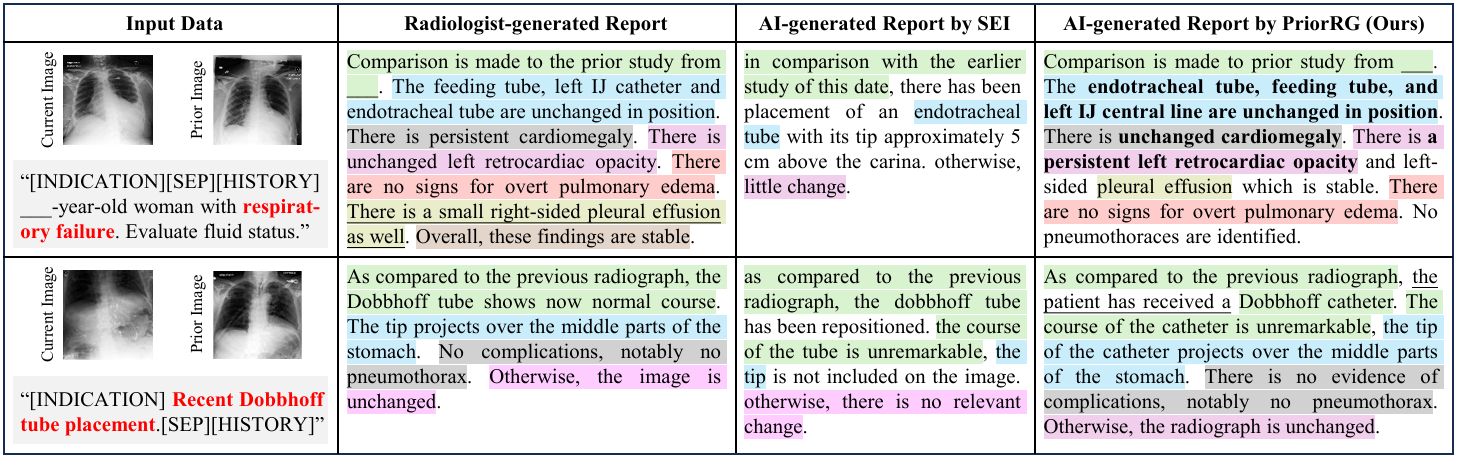}
    \caption{Qualitative comparison of reports from the baseline \cite{sei}, our proposed PriorRG, and radiologists. Sentences in the radiologist-generated report are color-coded to match corresponding descriptions in the AI-generated reports. In PriorRG's output, correct disease progression patterns and failure descriptions are marked in \textbf{bold} and \underline{underlined}, respectively.}
    \label{fig:qualitative-analysis}
\end{figure*}

\textbf{Effect of prior-aware coarse-to-fine decoding for report generation (Stage 2).} As shown in Table \ref{tab:ablation-study}, PriorRG significantly outperforms the baseline (a), which omits Stage 2, across both NLG and CE metrics, highlighting the critical role of Stage 2 in improving overall report quality.

\textbf{Effect of patient-specific prior knowledge.} Compared to variant (b) in Table \ref{tab:ablation-study}, which excludes patient-specific prior knowledge, PriorRG shows a marked improvement in both NLG and CE metrics. This result reveals the positive impact of such prior knowledge on chest X-ray report generation. Additional evidence can be found in Table~\ref {tab:breakdown}.

\textbf{Effect of the clinical context (CC).} In the image-text retrieval task (Figure~\ref{fig:retrieval}), PriorRG with CC outperforms its counterpart without CC, improving cross-modal alignment. In the report generation task, adding CC to variant (a) (see Table~\ref{tab:ablation-study}) yields consistent gains across both NLG and CE metrics. This result suggests that CC enhances the model's ability to capture individual information, improving the quality of generated reports.


\textbf{Effect of prior images (PI).} In the image-text retrieval task (Figure~\ref{fig:retrieval}), PriorRG with PI outperforms its variant without PI in the Stu-P@K metric, demonstrating its ability to model lesion progression through temporal image comparison. 
In the RRG task (Table~\ref{tab:ablation-study}), comparisons such as (c) vs. (e) and PriorRG vs. (d) show that incorporating PI consistently improves both NLG and CE metrics. These results highlight the value of prior images in enhancing study-level retrieval and improving report generation quality.


\textbf{Effect of hierarchical visual features.} PriorRG's gain over variant (e) in Table~\ref{tab:ablation-study} confirms that incorporating hierarchical visual features enhances report quality by capturing low-level details and high-level semantic cues.



\begin{table}
\centering
\setlength{\tabcolsep}{1.4mm}
\begin{tabular}{ccccccccc} 
\toprule
\multirow{2}{*}{\textbf{Model}} & \multicolumn{3}{c}{\textbf{NLG Metrics} $\uparrow$} & \multicolumn{3}{c}{\textbf{CE Metrics} $\uparrow$} \\ 
\cmidrule(lr){2-4}\cmidrule(r){5-7} & \textbf{B-2} & \textbf{B-4} & \textbf{MTR} & \textbf{P} & \textbf{R} & \textbf{F1} \\ 
\midrule
LastOnly & 0.278 & 0.168 & 0.185 & \textbf{0.554} & 0.466 & 0.506 \\
Fine2coarse & 0.284 & 0.172 & 0.187 & 0.549 & 0.482 & \textbf{0.514} \\
\textbf{PriorRG} & \textbf{0.290} & \textbf{0.175} & \textbf{0.189} & 0.541 & \textbf{0.485} & 0.511 \\
\bottomrule
\end{tabular}
\caption{Comparison of different progressive fusion strategies for report generation on the MIMIC-CXR dataset.}
\label{tab:progressive}
\end{table}

\begin{table}
    \centering
    \setlength{\tabcolsep}{1.1mm}
    \begin{tabular}{ccccccc}
    \toprule
    \textbf{Metric} & \textbf{w/ PI} & \textbf{w/o PI} & \textbf{w/ CC} & \textbf{w/o CC} & \textbf{w/ PK} & \textbf{w/o PK} \\
    \midrule
    B-2 $\uparrow$    & \textbf{0.289} & 0.288  & \textbf{0.294} & 0.139  & \textbf{0.294} & 0.140  \\
    F1 $\uparrow$    & \textbf{0.516} & 0.514  & \textbf{0.508} & 0.492  & \textbf{0.512} & 0.497  \\
    \bottomrule
    \end{tabular}
    \caption{Report generation performance on the MIMIC-CXR dataset, grouped by presence or absence CC (clinical context), PI (prior image), or PK (prior knowledge).}
    \label{tab:breakdown}
\end{table}

\textbf{Impact of progressive fusion strategies on report generation.} Table~\ref{tab:progressive} compares three progressive fusion strategies for the RRG task. The \textbf{LastOnly} variant utilizes only the final fused representation ${\boldsymbol{\bar{V}}}^{hier}$, discarding intermediate representations. The \textbf{Fine2coarse} variant is structurally identical to PriorRG, except that it reverses the order of spatiotemporal and hierarchical features integration. Results show that PriorRG consistently outperforms both variants across all NLG metrics, demonstrating the effectiveness of coarse-to-fine fusion. While Fine2coarse achieves a slightly higher F1-score, it falls behind PriorRG in linguistic fluency. The LastOnly variant performs worst, emphasizing the importance of retaining intermediate representations to support richer, more coherent report generation.


\textbf{Qualitative analysis.} Figure~\ref{fig:qualitative-analysis} presents a comparative visualization of generated reports on the MIMIC-CXR dataset. In AI-generated reports, more diverse colors indicate broader coverage of clinical findings, while longer bars reflect more accurate and detailed descriptions. The results show that: (1) PriorRG generates high-quality drafts requiring minimal revision---for example, in Case 1, only a brief note on \textit{pleural effusion} is added by the radiologist. (2) PriorRG captures disease progression effectively, as seen in the correct description of \textit{unchanged cardiomegaly} in Case 1. (3) PriorRG responds to clinical context---e.g., in Case 2, the model correctly addresses concern about \textit{Dobbhoff tube placement} and identifies potential complications. These examples demonstrate PriorRG's ability to track disease progression and understand clinical intent. Additional examples and failure case analysis are provided in Appendix B.

\section*{Conclusion}
We introduced \textbf{PriorRG}, a chest X-ray report generation framework that leverages clinical priors to enhance both image-text alignment and report quality. Specifically, we proposed a prior-guided contrastive pre-training scheme that mirrors diagnostic reasoning by using clinical context to guide spatiotemporal features extraction. Next, we presented a prior-aware coarse-to-fine decoding that progressively integrates clinical context, disease progression patterns, and hierarchical visual cues, enhancing the clinical accuracy and fluency of generated reports. Extensive experiments validated the effectiveness of our PriorRG on medical image-text retrieval and report generation. Future work will explore the organ-aware diagnosis framework \cite{Gu_2024_WACV-organ-mask} to further enhance interpretability.

\section*{Acknowledgments}
The work was jointly supported by the National Natural Science Foundations of China [grant number: 62272364], the Provincial Key Research and Development Program of Shaanxi [grant number: 2024GH-ZDXM-47], the Fundamental Research Funds for the Central Universities and the Innovation Fund of Xidian University [grant number: YJSJ25012].

\appendix

\begin{table}
\centering
\begin{tabular}{cccc} 
\toprule
\textbf{Prior Type} & \textbf{Ratio} & \textbf{B-2$\uparrow$} & \textbf{F1$\uparrow$} \\ 
\midrule
\multirow{5}{*}{Prior Image (PI)} 
 & 0\% & 0.287 & 0.513 \\
 & 10\% & 0.287 & 0.513 \\
 & 30\% & 0.288 & 0.513 \\
 & 50\% & \textbf{0.289} & \textbf{0.515} \\
 & 100\% & \textbf{0.289} & \textbf{0.515} \\
\cmidrule(r){1-4}
\multirow{5}{*}{Clinical Context (CC)} 
 & 0\% & 0.139 & 0.492 \\
 & 10\% & 0.157 & 0.495 \\
 & 30\% & 0.187 & 0.498 \\
 & 50\% & 0.220 & 0.501 \\
 & 100\% & \textbf{0.294} & \textbf{0.507} \\
\bottomrule
\end{tabular}
\caption{Report generation performance on the MIMIC-CXR dataset under varying proportions of Prior Image (PI) and Clinical Context (CC). The experiment moves beyond the binary setting of Table~\ref{tab:breakdown} by gradually adjusting the availability of prior knowledge. 
}
\label{atab:breakdown-var}
\end{table}

\section{Implementation Details}
\label{appendix-implementation-details}
\textbf{Model architecture.} The image encoder (RAD-DINO) \cite{2024-rad-dino} remains frozen throughout training. In contrast, the text encoder (CXR-BERT) \cite{2022-eccv-cxr-bert}, the report generator (DistilGPT2) \cite{Sanh2019DistilBERTAD}, projection heads, the attention-enhanced layer fusion network (ALF), and the spatiotemporal fusion network (STF) are fine-tuned during both training stages. All experiments are conducted on a single NVIDIA V100 GPU, with a memory usage of approximately 26 GB.

\textbf{Optimizer and scheduling.} We adopt the AdamW optimizer for all components. The ReduceLROnPlateau scheduler is employed with a patience of 5. Early stopping is applied with a patience of 15 epochs to prevent overfitting. 

\textbf{Training schedule.} We adopt a two-stage training pipeline. \textbf{For the MIMIC-CXR dataset} \cite{johnson-mimic-cxr-jpg}, the model is trained for 30 epochs in Stage 1 (prior-guided contrastive pre-training) with a batch size of 32, followed by another 30 epochs of fine-tuning in Stage 2 (prior-aware coarse-to-fine decoding) with a batch size of 16. In Stage 1, the learning rate is set to 5e-5. In Stage 2, the report generator is optimized with a learning rate of 5e-5, while all other components are trained with a learning rate of 5e-6. \textbf{For the MIMIC-ABN dataset} \cite{mimic-abn-ori}, as a derivative of MIMIC-CXR, the model is initialized with weights pre-trained on MIMIC-CXR and fine-tuned for 30 epochs using a learning rate of 5e-6 and a batch size of 16.

\textbf{Metrics implementation details.} All metrics are implemented using the pycocoevalcap \cite{chen2015microsoft-coco-caption} and f1chexbert \cite{Smit2020_chexbert} libraries. For all metrics, higher values indicate better performance.

\textbf{Medical image-text retrieval.} BioViL-T \cite{2023-cvpr-biovit-2301} and BiomedCLIP \cite{zhang-2023-biomedclip} are implemented using their officially released pre-trained weights and inference pipelines. Our PriorRG model is initialized with weights obtained from Stage 1 pre-training. For all three methods, the medical image-report retrieval is conducted by calculating feature similarity between image and report embeddings, followed by retrieving the top-K most similar reports.

\section{Additional Experiments}

\begin{table}
\centering
\setlength{\tabcolsep}{0.5mm}
\begin{tabular}{cccc} 
\toprule
\textbf{Method} & \textbf{Venu} & \textbf{BLEU-4$\uparrow$} & \textbf{METEOR$\uparrow$} \\ 
\midrule
\multicolumn{4}{l}{\textbf{\textit{Zero-shot  Report Generation}}} \\ 
R2GenGPT & Meta-Radio'23 & 0.112 & 0.177 \\
Med-LLM & MM'24 & 0.125 & 0.185 \\
\textbf{ PriorRG (Ours) } & - & \textbf{ 0.178 } & \textbf{ 0.211 } \\ 
\midrule
\multicolumn{4}{l}{\textbf{\textit{Supervised  Report Generation}}} \\ 
METransformer & CVPR'23 & 0.172 & 0.192 \\
MAN & AAAI'24 & 0.170 & 0.213 \\
DCG & MM'24 & 0.186 & 0.211 \\
Med-LLM & MM'24 & 0.168 & 0.209 \\
R2-LLM & AAAI'24 & 0.184 & 0.208 \\
\textbf{ PriorRG (Ours) } & - & \textbf{ 0.196 } & \textbf{ 0.216 } \\
\bottomrule
\end{tabular}
\caption{Comparison of report generation results on the IU X-ray dataset \cite{demner2016preparing-iu-xray} under zero-shot and supervised settings. \textbf{Best} results are in \textbf{Bold}.}
\label{tab:iu-xray}
\end{table}

\begin{table}
\centering
\setlength{\tabcolsep}{1.8mm}
\begin{tabular}{cccccc} 
\toprule
\multirow{2}{*}{Method} & \multirow{2}{*}{M/S} & \multicolumn{2}{c}{Stage
  1} & \multicolumn{2}{c}{Stage
  2} \\ 
\cmidrule(lr){3-4}\cmidrule(lr){5-6}
 &  & Trait & \#Param & Trait & \#Param \\ 
\midrule
MLRG & M & DP & 191M & DP+CC & 296M \\
\textbf{PriorRG} & S & DP+CC & 239M & DP+CC & 335M \\
\bottomrule
\end{tabular}
\caption{Comparison between our PriorRG and the recent baseline MLRG \cite{mlrg} in terms of input requirements and model size. ``M/S'' denotes multi-view or single-view current images; ``CC'' and ``DP'' indicate whether a method incorporates clinical context and disease progression, respectively. ``\#Param'' refers to the total number of model parameters.}
\label{tab:mlrg-input}
\end{table}

\begin{table*}
\centering
\begin{tabular}{lcccccc} 
\toprule
\textbf{Method} & \textbf{Stu-P@1} & \textbf{Stu-P@3} & \textbf{Stu-P@5} & \textbf{Cat-P@1} & \textbf{Cat-P@3} & \textbf{Cat-P@5} \\ 
\midrule
MLRG & 0.096 & 0.096 & 0.088 & 0.450 & 0.448 & 0.437 \\
PriorRG w/o PK & 0.279 & 0.238 & 0.191 & 0.601 & 0.574 & \textbf{0.549} \\
PriorRG w/o CC & 0.280 & 0.225 & 0.173 & 0.602 & 0.561 & 0.526 \\
PriorRG w/o PI & 0.393 & 0.324 & 0.245 & 0.637 & \textbf{0.597} & 0.548 \\
\textbf{PriorRG (Ours)} & \textbf{0.402} & \textbf{0.334} & \textbf{0.251} & \textbf{0.640} & 0.596 & 0.544 \\
\bottomrule
\end{tabular}
\caption{Medical image-text retrieval results on the MIMIC-5x200 dataset. PI: prior image; CC: clinical context; PK: prior knowledge. Cat-P@K measures whether retrieved reports belong to the same disease category as the query image, while Stu-P@K evaluates whether retrieved reports originate from the same study.}
\label{tab:mlrg-retrieval}
\end{table*}

\begin{table*}
\centering
\setlength{\tabcolsep}{1.4mm}
\begin{tabular}{cccccccccccccc} 
\toprule
\multirow{2}{*}{\textbf{ Method }} & \multirow{2}{*}{\textbf{ Venu }} & \multicolumn{6}{c}{\textbf{ NLG Metrics $\uparrow$}} & \multicolumn{4}{c}{\textbf{ CE Metrics $\uparrow$}} & \multirow{2}{*}{\makecell{\textbf{Time} \\ \textbf{(s/sample)}}$\downarrow$} & \multirow{2}{*}{\textbf{ GPU $\downarrow$}} \\ 
\cmidrule(lr){3-8}\cmidrule(lr){9-12}
 &  & \textbf{ B-1 } & \textbf{ B-2 } & \textbf{ B-3 } & \textbf{ B-4 } & \textbf{ MTR } & \textbf{ R-L } & \textbf{ P } & \textbf{ R } & \textbf{ F1 } & \textbf{ RG } &  &  \\ 
\midrule
MLRG & CVPR’25 & 0.411 & 0.277 & 0.204 & 0.158 & 0.176 & 0.320 & \textbf{0.549} & 0.468 & 0.505 & 0.293 & 0.2703 & 12.14GB \\
\textbf{PriorRG} & - & \textbf{0.412} & \textbf{0.290} & \textbf{0.220} & \textbf{0.175} & \textbf{0.189} & \textbf{0.324} & 0.541 & \textbf{0.485} & \textbf{0.511} & \textbf{0.296} & \textbf{0.1243} & \textbf{7.68 GB} \\

\bottomrule
\end{tabular}
\caption{Performance comparison of report generation on the MIMIC-CXR dataset. \textbf{Best} results are highlighted in \textbf{Bold}. RG denotes F1-RadGraph \cite{jain-radgraph} metric, which evaluates the overlap of clinical entities and relations, and better aligns with radiologists' assessments \cite{yu2023evaluating}.}
\label{tab:mlrg-performance}
\end{table*}

\subsection{Ablation Study on the Contribution of Prior Knowledge}

To quantify the contribution of different forms of prior knowledge, we perform an ablation study on the MIMIC-CXR dataset \cite{johnson-mimic-cxr-jpg} by varying the proportions of Prior Image (PI) and Clinical Context (CC) provided to the model. Unlike the binary on/off setup in Table~\ref{tab:breakdown}, we gradually adjust the availability of each type of prior knowledge to five levels: 0\%, 10\%, 30\%, 50\%, and 100\%. 

As shown in Table~\ref{atab:breakdown-var}, both BLEU-2 (B-2) and F1 consistently improve as more prior knowledge becomes accessible. PI shows only marginal influence, and PriorRG maintains stable performance even when PI is largely absent, highlighting its robustness to patient-specific sparsity. In contrast, CC yields a substantially stronger performance gain across all metrics, indicating that contextual clinical cues provide more actionable guidance for report generation. This trend aligns with radiological reasoning, where the integration of patient context is essential for accurate interpretation of imaging findings.

\subsection{Comparison with SOTA Methods on the IU X-ray Dataset in Zero-shot and Supervised Settings}
Following previous works \cite{chen-etal-2020-generating,chen-etal-2021-cross-modal}, we split the IU X-ray dataset \cite{demner2016preparing-iu-xray} into training, validation, and test sets with a 7:1:2 ratio. The detailed distribution is presented in Table~\ref{tab:statistic}.

We conduct experiments under both zero-shot and supervised settings, as illustrated in Table~\ref{tab:iu-xray}. In the zero-shot setting, the model trained during Stage 2 on the MIMIC-CXR dataset is directly applied to the IU X-ray test set without further fine-tuning. In the supervised setting, the model is initialized with the same MIMIC-CXR pretrained weights and then fine-tuned on the IU X-ray training set. The validation set is used for model selection, and final performance is reported on the test set.

Results in Table~\ref{tab:iu-xray} show the following: (1) In the zero-shot setting, PriorRG achieves the highest BLEU-4 (0.178) and METEOR (0.211), clearly outperforming baselines like Med-LLM \cite{liu2024in-context-acmmm} and R2GenGPT \cite{wang-2023-r2gengpt}. (2) In the supervised setting, PriorRG again achieves the best results, with BLEU-4 of 0.196 and METEOR of 0.216, outperforming strong baselines including METransformer \cite{wang2023metransformer}, MAN \cite{shen2024automatic_aaai}, DCG \cite{liang2024divide-acmmm-24}, Med-LLM \cite{liu2024in-context-acmmm}, and R2-LLM \cite{aaai-liu2024bootstrapping-llm}. (3) \textbf{These results demonstrate the strong generalization capability of our PriorRG across datasets and its robustness, even when prior images are unavailable.}

\subsection{Comparison with Concurrent Work MLRG on the MIMIC-CXR Dataset}
MLRG \cite{mlrg}, a concurrent work accepted at CVPR 2025, also jointly models temporal visual changes and clinical context. To enable a comprehensive comparison, we evaluate both methods in terms of methodological design (Table~\ref{tab:mlrg-input}), medical image-text retrieval (Table~\ref{tab:mlrg-retrieval}), and report generation (Table~\ref{tab:mlrg-performance}). \textbf{(1) Methodology:} Unlike MLRG \cite{mlrg}, our PriorRG explicitly models disease progression and clinical context during cross-modal alignment (Stage 1; see Table~\ref{tab:mlrg-input}). This design leads to significantly improved retrieval performance, as reflected by higher Stu-P@K and Cat-P@K scores (Table~\ref{tab:mlrg-retrieval}). \textbf{(2) Report Generation:} On the chest X-ray report generation task, PriorRG consistently outperforms MLRG across natural language generation (NLG) and clinical efficacy (CE) metrics, achieving a +1.7\% improvement in BLEU-4 (B-4) and +0.6\% in F1-CheXbert (F1). \textbf{(3) Efficiency and Deployment:} PriorRG requires only a single-view current image while delivering higher-quality reports. It achieves over 2× faster inference and reduces GPU memory usage by nearly 5 GB. All comparisons are conducted on the MIMIC-CXR test set (3,852 samples) under identical hardware, batch size, and num\_worker configurations, demonstrating PriorRG's practical efficiency and deployment readiness.

\begin{figure*}
  \includegraphics[width=\textwidth]{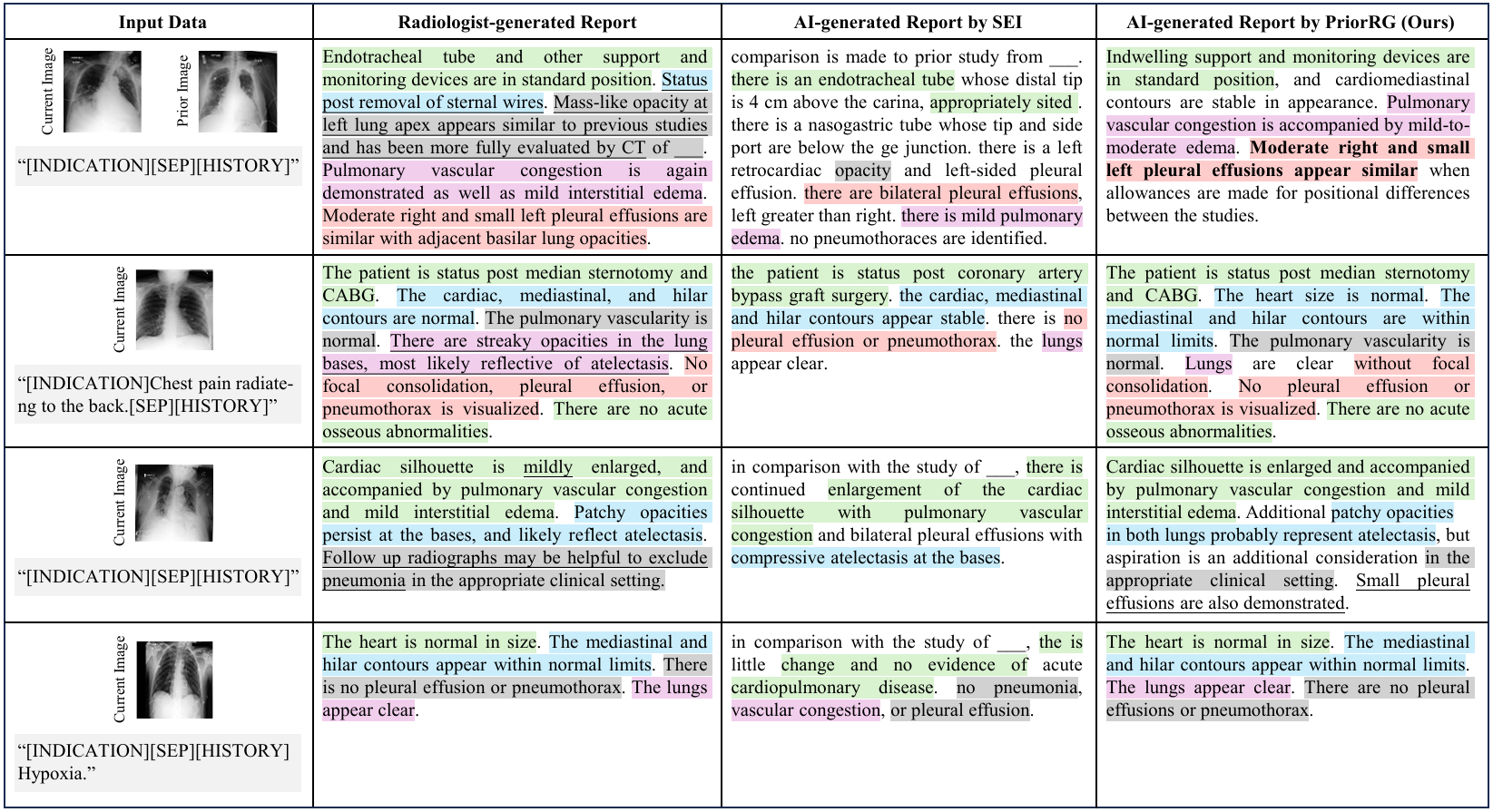}
  \caption{Qualitative comparison of reports from the baseline \cite{sei}, our proposed PriorRG, and radiologists. Sentences in the radiologist-generated report are color-coded to match corresponding descriptions in the AI-generated reports. In PriorRG's output, correct disease progression patterns and failure descriptions are marked in \textbf{bold} and \underline{underlined}, respectively.}
  \label{fig:teaser}
\end{figure*}

\begin{table*}
\centering
\begin{tabular}{cccccccc} 
\toprule
\multirow{2}{*}{\textbf{Observation}} & \multirow{2}{*}{\textbf{\%}} & \multicolumn{3}{c}{\textbf{SEI} \cite{sei}} & \multicolumn{3}{c}{\textbf{PriorRG (Ours)}} \\ 
\cmidrule(lr){3-5}\cmidrule(lr){6-8}
 &  & \textbf{Precision} & \textbf{Recall} & \textbf{F1-score} & \textbf{Precision} & \textbf{Recall} & \textbf{F1-score} \\ 
\midrule
Enlarged Cardiomediastinum & 9.9 & \textbf{0.373} & 0.208 & 0.267 & 0.361 & \textbf{0.399} & \textbf{0.379} \\
Cardiomegaly & 14.8 & 0.599 & 0.633 & 0.616 & \textbf{0.615} & \textbf{0.687} & \textbf{0.649} \\
Lung Opacity & 13.8 & 0.519 & 0.170 & 0.256 & \textbf{0.581} & \textbf{0.338} & \textbf{0.427} \\
Lung Lesion & 2.5 & \textbf{0.462} & 0.021 & 0.041 & 0.286 & \textbf{0.061} & \textbf{0.100} \\
Edema & 8.3 & \textbf{0.526} & 0.361 & 0.428 & 0.468 & \textbf{0.471} & \textbf{0.469} \\
Consolidation & 3.3 & 0.218 & \textbf{0.194} & 0.205 & \textbf{0.341} & 0.171 & \textbf{0.228} \\
Pneumonia & 4.4 & 0.174 & 0.065 & 0.095 & \textbf{0.282} & \textbf{0.214} & \textbf{0.243} \\
Atelectasis & 10.9 & 0.469 & 0.395 & 0.429 & \textbf{0.486} & \textbf{0.437} & \textbf{0.460} \\
Pneumothorax & 1.0 & 0.174 & 0.039 & 0.064 & \textbf{0.381} & \textbf{0.320} & \textbf{0.348} \\
Pleural Effusion & 12.4 & 0.683 & \textbf{0.697} & \textbf{0.690} & \textbf{0.722} & 0.630 & 0.672 \\
Pleural Other & 1.6 & 0.167 & 0.022 & 0.039 & \textbf{0.203} & \textbf{0.071} & \textbf{0.106} \\
Fracture & 1.8 & 0.000 & 0.000 & 0.000 & \textbf{0.250} & \textbf{0.043} & \textbf{0.074} \\
Support Devices & 12.8 & 0.763 & 0.708 & 0.734 & \textbf{0.768} & \textbf{0.755} & \textbf{0.761} \\
No Finding & 2.4 & 0.161 & \textbf{0.597} & 0.253 & \textbf{0.257} & 0.561 & \textbf{0.353} \\ 
\cmidrule(lr){1-8}
micro avg & - & 0.523 & 0.410 & 0.460 & \textbf{0.541} & \textbf{0.485} & \textbf{0.511} \\
macro avg & - & 0.378 & 0.294 & 0.294 & \textbf{0.429} & \textbf{0.368} & \textbf{0.376} \\
\bottomrule
\end{tabular}
\caption{Clinical accuracy of 14 observations on the MIMIC-CXR dataset. Best values are highlighted in bold.}
\label{tab:chexbert-details}
\end{table*}

\begin{table*}
\centering
\begin{tabular}{cccccccccc} 
\toprule
\multirow{2}{*}{\textbf{ Statistic}} & \multicolumn{3}{c}{\textbf{ MIMIC-CXR }} & \multicolumn{3}{c}{\textbf{ MIMIC-ABN }} & \multicolumn{3}{c}{\textbf{ IU X-ray }} \\ 
\cmidrule(lr){2-4}\cmidrule(lr){5-7}\cmidrule(lr){8-10}
 & \textbf{ Train } & \textbf{ Val } & \textbf{ Test } & \textbf{ Train } & \textbf{ Val } & \textbf{ Test } & \textbf{ Train } & \textbf{ Val } & \textbf{ Test } \\ 
\midrule
\#Image & 239,998 & 2,113 & 3,852 & 69,641 & 586 & 844 & 4,138 & 592 & 1,180 \\ 
\#Sample & 239,998 & 2,113 & 3,852 & 69,641 & 586 & 844 & 2,069 & 296 & 590 \\ 
\#Report & 150,957 & 1,182 & 2,343 & 34,763 & 263 & 378 & 2,069 & 296 & 590 \\ 
\%PI & 60.5 & 61.3 & 87.7 & 52.7 & 50.6 & 89.2 & 0.0 & 0.0 & 0.0 \\ 
\%Indication & 66.4 & 65.4 & 57.3 & 64.6 & 62.7 & 56.3 & 87.4 & 86.1 & 85.6 \\ 
\%History & 30.6 & 31.5 & 37.4 & 34.9 & 37.3 & 43.4 & 0.0 & 0.0 & 0.0 \\
\bottomrule
\end{tabular}
\caption{Statistics of the MIMIC-CXR \cite{johnson-mimic-cxr-jpg}, MIMIC-ABN \cite{mimic-abn-ori}, and IU X-ray \cite{demner2016preparing-iu-xray} for training, validation, and test sets. ``\#Image'', ``\#Sample'', and ``\#Report'' denote the number of medical images, samples, and radiology reports, respectively. ``\%PI'', ``\%Indication'', and ``\%History'' represent the proportions of prior images, \textit{indications}, and medical \textit{history}, respectively.}
\label{tab:statistic}
\end{table*}


\subsection{Qualitative Analysis}
\label{sec:a-qualitative-analysis}
Additional qualitative analysis is shown in Figure~\ref{fig:teaser}. The results show that PriorRG outperforms the baseline \cite{sei} in generating context-aware, disease progression-sensitive radiology reports. \textbf{In cases with the prior image but no clinical context (Case 1)}, the model effectively captures disease progression by describing pleural effusions as stable over time. \textbf{When only clinical context is available}---such as \textit{chest pain radiating to the back} (Case 2) or \textit{hypoxia} (Case 4)---PriorRG generates radiology reports that align with the underlying clinical intent, accurately reflecting surgical history (e.g., \textit{CABG} in Case 2) and comprehensively assessing relevant cardiopulmonary structures in response to \textit{hypoxia} (Case 4). \textbf{Even in the absence of both the prior image and clinical context (Case 3)}, our PriorRG still produces complete and clinically plausible reports.

\subsection{Failure Case Analysis}
\label{sec:fail}
Figure~\ref{fig:teaser} illustrates that our PriorRG occasionally produces overstatements, such as simplifying \textit{mildly enlarged cardiac silhouette} to \textit{enlarged} (Case 3). This phenomenon mainly stems from the model's lack of effective modeling of fine-grained medical attributes, leading to insufficient control over the severity and semantic accuracy in generated reports. To address this limitation, we are exploring attributed abnormality graphs \cite{tmi-2024-scene-graph-rrg,yan_tmi_2023} to more accurately capture and represent attribute-specific disease states.

\bibliography{aaai2026}

\end{document}